\documentclass{article}

\usepackage[final,position]{neurips_2026}

\usepackage[utf8]{inputenc} % allow utf-8 input
\usepackage[T1]{fontenc}    % use 8-bit T1 fonts
\usepackage{hyperref}       % hyperlinks
\usepackage{url}            % simple URL typesetting
\usepackage{booktabs}       % professional-quality tables
\usepackage{amsfonts}       % blackboard math symbols
\usepackage{nicefrac}       % compact symbols for 1/2, etc.
\usepackage{microtype}      % microtypography
\usepackage{xcolor}         % colors
\usepackage{comment}
\usepackage{graphicx}
\usepackage[breakable,skins]{tcolorbox}
\newtcolorbox[auto counter]{casebox}[2][]{%
  enhanced, breakable,
  colback=gray!5, colframe=gray!55,
  boxrule=0.5pt, arc=1pt,
  left=6pt, right=6pt, top=4pt, bottom=4pt,
  colbacktitle=gray!18, coltitle=black,
  fonttitle=\bfseries,
  before upper={\setlength{\parskip}{5.5pt}\setlength{\leftmargini}{1.5em}},
  title={Box~\thetcbcounter: #2},
  #1}

\title{Mechanistic Interpretability Needs Philosophy}

\author{
  Iwan Williams$^1$ \And
  Ninell Oldenburg$^1$ \And
  Ruchira Dhar$^2$ \And
  Joshua Hatherley$^{3~\dagger}$ \And
  Constanza Fierro$^2$ \And
  Nina Rajcic$^1$ \And
  Sandrine R. Schiller$^\dagger$ \And
  Filippos Stamatiou$^1$ \And
  Anders Søgaard$^{1,2}$ \\[1em]
  $^1$University of Copenhagen, Center for Philosophy of Artificial Intelligence,\\ Section for Philosophy, Department of Communication\\
  $^2$University of Copenhagen, Department of Computer Science\\
  $^3$Macquarie University, Department of Philosophy\\
}

\begin{document}

\maketitle
{
  \renewcommand{\thefootnote}{\ensuremath{\dagger}}
  \footnotetext{Work conducted at the Center for Philosophy of Artificial Intelligence, University of Copenhagen.}
}

\begin{abstract}
Mechanistic interpretability (MI) aims to explain how trained neural networks work by uncovering their underlying mechanisms. As the field grows in influence, it is increasingly important to examine not just models themselves, but the assumptions, concepts and explanatory strategies implicit in MI research. We argue that \textit{mechanistic interpretability needs philosophy} as an ongoing partner in clarifying its concepts, refining its methods, and navigating the epistemic and ethical complexities of interpreting AI systems. There is significant unrealised potential for progress in MI to be gained through deeper engagement with philosophers and philosophical frameworks. Taking three open problems from the MI literature as examples, this paper illustrates the value philosophy can add to MI research, and outlines a path toward deeper interdisciplinary dialogue.
\end{abstract}

%%%%%%%%%%%%%%%%%%%%%%%%%%%%%%%%%%%%%%%%%%%%%%%%%%%%%%%%%%%%%%%%%%%%%%%
%%%%%%%%%%%%%%%%%%%%%%%%%%%%%%%%%%%%%%%%%%%%%%%%%%%%%%%%%%%%%%%%%%%%%%%
%%%%%%%%%%%%%%%%%%%%%%%%%%%%%%%%%%%%%%%%%%%%%%%%%%%%%%%%%%%%%%%%%%%%%%%
%%%%%%%%%%%%%%%%%%%%%%%%%%%%%%%%%%%%%%%%%%%%%%%%%%%%%%%%%%%%%%%%%%%%%%%
%%%%%%%%%%%%%%%%%%%%%%%%%%%%%%%%%%%%%%%%%%%%%%%%%%%%%%%%%%%%%%%%%%%%%%%
%%%%%%%%%%%%%%%%%%%%%%%%%%%%%%%%%%%%%%%%%%%%%%%%%%%%%%%%%%%%%%%%%%%%%%%

\section{Introduction}\label{sec:introduction}

How and why do artificial neural networks produce the outputs they do? Since the resurgence of deep learning approximately a decade ago, this question has driven various efforts to interpret and explain AI systems. Among these efforts, mechanistic interpretability (MI) has emerged as an increasingly influential strand of research \citep{saphra2024mechanistic}. Yet, despite its rapid rise, MI is often described as a ``pre-paradigmatic'' field \citep{bereska2024mechanistic}: several foundational open problems remain unsolved \citep{sharkey2025open} and MI has faced significant critiques \citep{adolfi2024complexity, meloux2025everything, makelov2024is}. Making progress on these problems will require input from various perspectives and skill sets. In this paper, we focus on a cross-disciplinary contribution whose full potential has been overlooked: we argue that \textbf{mechanistic interpretability needs philosophy}—philosophy can make a crucial contribution to progress in MI. While some MI researchers have begun to acknowledge the existence of ``philosophical'' questions in their field \citep{sharkey2025open, fierro2024defining}, the potential role of philosophy in MI remains underappreciated. MI can make greater and more efficient progress towards its scientific and societal goals through increased dialogue with philosophers and philosophical frameworks. To make our case, we examine three broad research questions in MI and show how resources from philosophy can facilitate progress on each of them. The examples we discuss serve to illustrate the various ways philosophy can add value to MI research. Our position parallels arguments that have been made for the crucial role of philosophy in the fields of physics \citep{rovelli2018physics}, cognitive science \citep{thagard2009cognitive}, economics \citep{nussbaum2016economics}, AI research more broadly \citep{buckner2024deep}, and science in general \citep{laplane2019science}.

\paragraph{What is mechanistic interpretability?}\label{sec:whatismi}
Artificial neural networks are notoriously opaque. \textit{AI Interpretability} \citep{ghosh2020interpretable, sathyan2022interpretable, shin2022platforms, calderon2024behalf} and \textit{Explainable Artificial Intelligence (XAI)} \citep{goebel2018explainable, miller2019explanation, xu2019explainable, dwivedi2023explainable} encompass various approaches to address the issue of opacity.\footnote{We will use these terms interchangeably, though we note that others use ``interpretability'' and ``explainability'' to delineate two different research agendas \citep{miller2019explanation}.} 
While there are many XAI taxonomies \citep{miller2019explanation,guidotti2018survey,carvalho2019machine,molnar2020interpretable,kotonya2020explainable,sogaard2021explainable}, one central distinction is between attempts to create models that are interpretable by design and attempts to interpret ``black box'' models \citep{rudin2019stop, lakkaraju2019faithful}. Within this latter branch, we can distinguish between the goal of generating explanations for various \textit{non-specialist audiences}  
and that of developing explanations for \textit{theorists} \citep{chalmers2025propositional}. The latter goal requires emulating something like the scientific method, i.e., an iterative, coordinated research strategy in which a range of experimental methods are deployed, data are integrated, hypotheses are tested, and theories are refined \citep{kastner2024explaining}. 
It is on this branch of the XAI tree that we find mechanistic interpretability. While there are various kinds of explanation that theorists might aim for, \textit{mechanistic} interpretability (as the name suggests) is characterised by the search for the \textit{mechanisms} by which a model works \citep{sharkey2025open, bereska2024mechanistic}. We discuss mechanistic explanations in detail in Section \ref{decomposition}.\footnote{While ``mechanistic interpretability'' is sometimes used narrowly to refer to a specific, culturally identified research community \citep{saphra2024mechanistic} (e.g. to those associated with Distill.pub’s \textit{Circuits} thread \citep{olah2020zoom} and Anthropic’s \textit{Transformer Circuits} thread \citep{elhage2022toy,olsson2022context}), we adopt a broader usage that includes (e.g.) prior and parallel work in the NLP and computer vision communities.}

\begin{figure}
    \centering
    \includegraphics[width=0.9\linewidth]{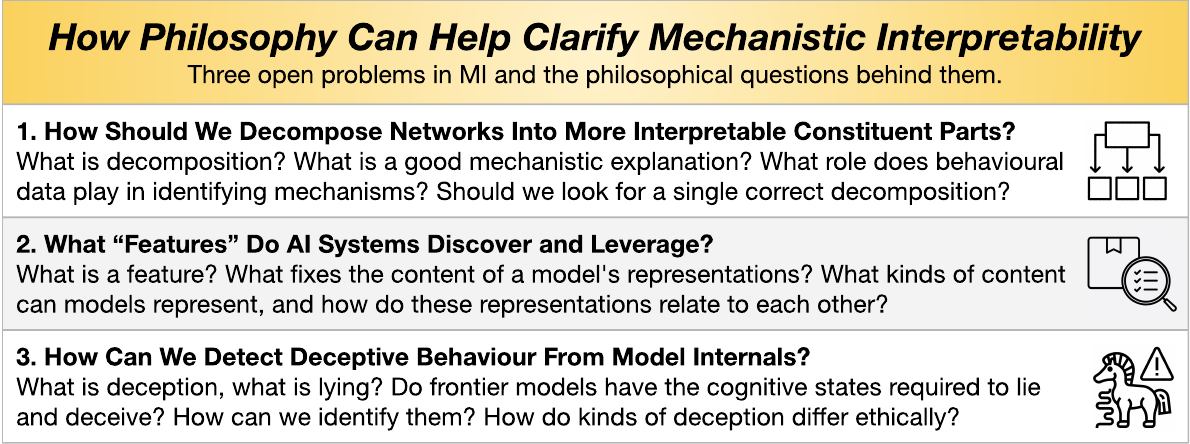}
    \caption{How philosophy can help: a case based on three open problems in MI.}
    \label{fig:mineedsphilosophy}
\end{figure}

\paragraph{Paper structure} Section \ref{Open} introduces three open problems in MI and highlights specific ways that philosophy can help to make progress on those problems (see Figure \ref{fig:mineedsphilosophy}). Section \ref{decomposition} discusses the challenge of decomposing networks into more interpretable constituent parts; Section \ref{sec:features} discusses the challenge of identifying the ``features'' leveraged by AI models; and Section \ref{deception} discusses the challenge of using MI methods to detect and mitigate deceptive behaviour by AI models. Section \ref{sec:objections} responds to objections to our claim that MI needs philosophy and offers practical recommendations for increasing fruitful interaction between the two fields.

%%%%%%%%%%%%%%%%%%%%%%%%%%%%%%%%%%%%%%%%%%%%%%%%%%%%%%%%%%%%%%%%%%%%%
%%%%%%%%%%%%%%%%%%%%%%%%%%%%%%%%%%%%%%%%%%%%%%%%%%%%%%%%%%%%%%%%%%%%%
%%%%%%%%%%%%%%%%%%%%%%%%%%%%%%%%%%%%%%%%%%%%%%%%%%%%%%%%%%%%%%%%%%%%%
%%%%%%%%%%%%%%%%%%%%%%%%%%%%%%%%%%%%%%%%%%%%%%%%%%%%%%%%%%%%%%%%%%%%%
%%%%%%%%%%%%%%%%%%%%%%%%%%%%%%%%%%%%%%%%%%%%%%%%%%%%%%%%%%%%%%%%%%%%%
%%%%%%%%%%%%%%%%%%%%%%%%%%%%%%%%%%%%%%%%%%%%%%%%%%%%%%%%%%%%%%%%%%%%%

%%%%%%%%%%%%%%%%%%%%%%%%%%%%%%%%%%%%%%%%%%%%%%%%%%%%%%%%%%%%%%%%%%%%%
%%%%%%%%%%%%%%%%%%%%%%%%%%%%%%%%%%%%%%%%%%%%%%%%%%%%%%%%%%%%%%%%%%%%%
%%%%%%%%%%%%%%%%%%%%%%%%%%%%%%%%%%%%%%%%%%%%%%%%%%%%%%%%%%%%%%%%%%%%%
%%%%%%%%%%%%%%%%%%%%%%%%%%%%%%%%%%%%%%%%%%%%%%%%%%%%%%%%%%%%%%%%%%%%%
%%%%%%%%%%%%%%%%%%%%%%%%%%%%%%%%%%%%%%%%%%%%%%%%%%%%%%%%%%%%%%%%%%%%%
%%%%%%%%%%%%%%%%%%%%%%%%%%%%%%%%%%%%%%%%%%%%%%%%%%%%%%%%%%%%%%%%%%%%%

\section{Open problems in MI – and how philosophy can help}\label{Open}

We focus on three open problems in MI, drawing on \citet{sharkey2025open}, and show how philosophy can help investigate each (see Figure~\ref{fig:mineedsphilosophy}). As we shall see, philosophy's potential contributions are not solely critical or negative, i.e. identifying mistakes or confusions in existing approaches (though philosophy can, and has, served that important role, too). Engagement with philosophy can also be \textit{generative}, suggesting new avenues for empirical investigation, or novel interpretations of results.

\subsection{How should we decompose networks into more interpretable constituent parts?}\label{decomposition}

A core open problem in MI is how best to decompose complex neural networks into more interpretable constituent parts \citep{mueller2024quest, sharkey2025open}. Obvious structural components of neural networks, like neurons, parameters and attention heads, often fail to map cleanly onto functionally meaningful roles. As a result, interpretability researchers increasingly seek more abstract, distributed, or coarse-grained decompositions that better capture a model’s internal logic and behaviour \citep{bricken2023monosemanticity, shafran2025decomposing}.
While this is often framed as a purely empirical task, empirical work may be guided by unexamined assumptions about the criteria for a good explanation. Here, philosophy of science can offer useful frameworks and insights. As others have argued \citep{kastner2024explaining}, the field of MI ultimately seeks what philosophers call \textit{mechanistic explanations}—accounts of how phenomena arise from organized causal interactions among parts \citep{machamer2000thinking, craver2007explaining, bechtel2005explanation}.\footnote{Older work in XAI has had similar aspirations \citep{balkir-etal-2022-necessity}; we happily extend the label of MI to include this work, too.} Common across fields like biology, physics, cognitive science, and economics, explanations of this sort identify a \textit{mechanism}: a set of \textit{entities} and \textit{activities}, \textit{organised} to produce or maintain a phenomenon \citep{Glennan2022}. Mechanistic explanations do not just describe regularities but show \textit{how} they emerge from causal structure.\footnote{Thus, philosophers sometimes contrast mechanistic models with merely \textit{descriptive} (or ``phenomenal'') models, which don’t aim to capture the causal structure of a system \citep{kaplan2011explanatory}.} Thus, a key virtue of such explanations is that they support \textit{intervention}: by revealing the components and activities responsible for a phenomenon, they clarify how it might be changed or controlled \citep{woodward2005making,craver2013search}. The philosophical literature on mechanistic explanation provides a number of insights that can guide attempts to decompose networks into interpretable components, two of which we highlight below.

\paragraph{Emphasising the interrelation of mechanism and behaviour}

Mechanistic interpretability is often presented as targeting the ``inner'' workings of neural networks, in contrast to approaches that focus on outwardly observable behaviour \citep{rauker2023toward, 10.5555/3692070.3694093,grzankowski2024real}. Indeed, a common motivation for MI is the idea that, for any given task, there are many possible algorithms or solutions \citep[see, e.g.][]{zhong2023clock}, and thus a model's ability to achieve some input–output function does not tell us \textit{how} the function is carried out. It is thus tempting to dismiss ``behavioural'' approaches to studying AI systems as being distinct from (and perhaps inferior to) ``mechanistic'' approaches.\footnote{For instance, \citep{bereska2024mechanistic} approvingly gloss the move from “black-box” to “inner” interpretability as “akin to the evolution from behaviorism to cognitive neuroscience” (p. 1).} However, from the perspective of philosophical accounts of mechanistic explanation, this dichotomy between mechanistic and behavioural approaches is, at best, misleading. To be sure, MI already incorporates behavioural measures in at least one way: researchers often infer the role of internal components by intervening on them (e.g. through ablation, patching or steering) and observing the downstream effect on models' behaviour. Our point here is that MI has overlooked the full variety of ways that behaviour—including more purely behavioural methods—can inform the search for mechanisms.\footnote{See \citep{holtzman2026prompting} for a convergent perspective.} The various roles of behaviour come into view more easily in light of philosophical accounts of mechanistic explanation, which have long emphasised the need to look ``down, around and up'' to uncover causal mechanisms \citep{Bechtel2009}.

First, the process of mechanism discovery, as emphasised in many philosophical accounts, typically begins with identifying and characterising \textit{phenomena}—events, behaviours or conditions that call for explanation [\citealt[ch. 4]{craver2013search}, \citealt{dresow2025interdisciplinary}]. In the context of MI, investigations of model behaviour can (i) surface new targets for explanation, and (ii) systematically catalogue aspects of those phenomena that need explaining. Second, once a phenomena is identified, behaviour continues to be an important source of data that can inform and constrain hypotheses about the mechanisms that produce it. Philosophers have stressed the indispensability of behavioural data in formulating plausible models, or suggesting ``sketches'' for mechanisms \citep{piccinini2011integrating} and the same approach can fruitfully extend to mechanistic interpretability \citep{budding2024does}. In this context, studying behaviour should not be construed narrowly as benchmarking a model's success at a task. A more fruitful approach, relevant to the explanatory goals of MI, involves carefully designed studies that systematically map unexpected behaviours in edge cases, identify patterns of breakdown, and test for other behavioural ``signatures'' of specific algorithms \citep{taylor2022signature}.\footnote{For examples, see some recent work on Theory of Mind \citep{ullman2023large,strachan2024testing} and reasoning \citep{nezhurina2024alice, lewis2024evaluating} capacities in LLMs.} These insights can help refocus the efforts of MI, ensuring the field capitalises on all sources of data relevant to understanding neural networks mechanistically.

\paragraph{Challenging the assumption of the One True Decomposition}

Discussions of decomposition in MI are often framed as the search for the \textit{right} level of analysis, implying a privileged cut through a network that reveals its ``true'' structure. \citet{bereska2024mechanistic} describe the goal of much MI work as searching for ``the true, disentangled features'' (p. 8) or the ``ground truth network features'' (p. 16). Drawing on a Platonic metaphor, \citet{sharkey2025open} describe this aspiration as ``carving neural networks at their joints'' (p. 13). But this metaphor, while evocative, imports a unexamined assumption: that models have a unique and natural decomposition, independent of explanatory context.

This is often more than mere rhetoric and has direct bearing on the field’s interpretation of anomalies. For example, the use of sparse autoencoders (SAEs) to attempt to identify internal representations in LLMs (discussed further in Section \ref{sec:features}) has encountered complications such as “feature splitting” \citep{bricken2023monosemanticity}: based on experimenter choices, researchers find greater or fewer representational primitives, varying in their granularity, but lack a principled basis for deciding between the different decompositions. Feature splitting, and similar findings, are often taken to be problematic, on the assumption that there is a unique ``ground truth'' decomposition yet to be found. But, as we shall see, this assumption is open to question.

Philosophers of science have long emphasized that mechanisms span multiple levels of organization \citep{craver2014levels}, and mechanisms in biology, neuroscience, and cognitive science rarely submit to a single, correct level of analysis. In the life sciences, scientists might study whole ecosystems, individual organisms, systems of organs, mechanisms within cells, or molecular interactions. One view of how these perspectives fit together is in terms of mechanisms nested within mechanisms—what we treat as a simple activity of a component at one level (e.g. a neuron firing) can be subjected to a further ``how does it work?'' question, which can often be answered in terms of a lower-level mechanism (e.g. opening and closing ion channels). Importantly, no single level of description has a unique claim to being mechanistic—there are simply different levels of mechanisms. And taking a mechanistic approach need not involve treating lower-level mechanisms as more important (or more ``real'') than higher-level ones.\footnote{Thus, MI practitioners are on firm footing in dismissing the charge that their field is ``reductionist'' \citep{Hendrycks_Hiscott_2025}.}

In practice, which level of mechanism is most important depends largely on the pragmatic goals of researchers.\footnote{To be sure, MI researchers sometimes appear to have a more pluralist attitude (recognising pragmatic considerations for the utility of alternative decompositions). The value of philosophy is in making these meta-theoretical assumptions explicit, and thus helping to distinguish real empirical roadblocks from mere artifacts of explanatory schemes.} 
This insight has direct implications for MI: 
different decompositions may be more or less useful, whether we are trying to (say) control outputs, understand generalisation, or detect deception. 
One philosophical framework for thinking about this is \textit{explanatory pluralism}—the view that scientific understanding often requires integrating multiple, cross-cutting models, tailored to different explanatory aims \citep{mitchell2023landscape}.\footnote{Recent developments in MI echo this pluralist perspective. The \textit{causal abstraction} framework \citep{geiger2024finding, geiger2021causal, geiger2025causal} formalises the idea that high-level and low-level descriptions of neural networks can serve different explanatory goals. (It is perhaps no coincidence that this work is partly influenced by formal philosophical work on causation and explanation.)} Taken together, these philosophical insights can help to reframe and guide efforts to decompose neural networks into components. Decompositions should be evaluated not by how well they mirror the unique ``real'' structure of the network, but by how effectively they support causal understanding, prediction, and intervention across different research contexts.\footnote{Another valuable resource from the philosophy of science can be found in attempts to abstract, systematise, and codify mechanism-discovery strategies often implicit in scientific practices \citep{darden2006reasoning, craver2013search, darden2017strategies}. These accounts can help MI researchers to better understand their own toolkits and perhaps inspire new methods through analogies with other fields.}

\subsection{What “features” do AI systems discover and leverage?}
\label{sec:features}

While there are many kinds of natural and human-engineered mechanisms, a core, if not universal, working assumption in MI is that the fundamental components of deep neural networks are \emph{features}. Unpacking, clarifying, and interrogating this assumption are further ways philosophers can make a practical contribution to MI.

At a first pass, features involve a correspondence between an aspect of model internals and an external condition that the model leverages in carrying out some task. Early work in image classification models suggested that individual units (neurons) within a network may encode visual attributes, such as \textit{red}, or \textit{vertical edge}~\citep{zeiler2014visualizing}. But as models became larger and network architectures more complex, this straightforward mapping between neuron and feature has been questioned, with many MI researchers now hypothesising that features are encoded ``in superposition'' across multiple neurons—e.g. as almost-orthogonal directions in activation space \citep{elhage2022toy}. 

To explain neural networks in terms of features is, in philosophical jargon, to adopt a \textit{representational} lens. Representations, in the philosophical sense, are system-internal components whose function is to encode information (or ``carry content'') about things external to the system, and thus to drive appropriate behaviour. Feature-based explanations are representational explanations, because they are attempts to explain system behaviour in terms of internal representations, in this sense.\footnote{In the ML literature, ``representation'' is sometimes used in a looser sense to refer to any intermediate activation pattern (irrespective of whether it has a clearly defined content-encoding role). Presumably, though, the term representation is used precisely because they are assumed to play a representational function in the narrower sense (i.e., exploitable encoding of information). We will adopt this narrower usage in what follows.} The nature of representations—especially in the context of human and animal cognition—has long been discussed in philosophy, and this remains an active area of research. (For helpful introductions and book-length treatments, see \citep{ryder2009problems1,ryder2009problems2,shea2018representation,schulte2023mental}). Here we highlight three ways the philosophy of representation can contribute to MI research into features.

\paragraph{Distinguishing vehicles from content}
The term ``feature'' is often used in inconsistent ways, sometimes referring to an internal component of a model—such as a neuron, or a non-basis direction in activation space—and sometimes referring to a property of an input—such as a \textit{curve}, \textit{the Golden Gate Bridge} or \textit{positive sentiment}. \citet{Williams_2024} argues that this confusion rests on an equivocation between two aspects of a representation, which it has become standard practice to distinguish in the philosophical literature, namely the \textit{vehicle} and the \textit{content} of a representation. A representational vehicle is the internal symbol, signal, or aspect of activity whose function is to encode content (to detect, represent, or refer to something external). Representational vehicles enter into causal-computational relations with other representational vehicles and ultimately generate the behavioural output of a system. By contrast, the content of a representation is the task-relevant external condition (object, property, category, relation, proposition) that is represented by a representational vehicle, and which makes sense of the cascade of causal interactions between vehicles.\footnote{Philosophers have contributed to discussions over how best to individuate representational vehicles in neural networks for some time \citep{clark1993associative,shea2007content,azhar2016polytopes}, in many ways prefiguring recent debates about the linear representation hypothesis \citep{park2023linear} in mechanistic interpretability.} 

One benefit of clearly differentiating the vehicle sense of ``feature'' from the content sense of ``feature'' (beyond avoiding unnecessary confusion and cross-talk) is that it allows researchers to clearly articulate distinct research questions. For example, one line of inquiry is \textit{what contents do models learn to represent?} For instance, do language models represent causal information about real-world entities, or do they simply represent syntactic and distributional properties of words? Does a given model represent the states and properties of users? If so, \textit{which} properties? And so on. A quite different line of inquiry is \textit{what are the vehicles of content in ML models?} For instance, when do models develop representational vehicles that align with non-basis directions vs. with individual neurons? How do simple representational vehicles (e.g., those representing objects and properties) combine into complex representational vehicles (e.g., those representing facts)? And what are the \textit{specific} vehicles for \textit{specific} contents—e.g., the vehicle responsible for encoding the language of an input? 

\paragraph{Refining experimental approaches}
A key concern in philosophical discussions of representation has been the search for the right \textit{theory of content}. This is the question of what \textit{grounds} representational content—what makes it the case that a given representation represents \textit{cats} rather than \textit{dogs} or \textit{cat-shaped things}? In the context of AI models, we might similarly ask, what properties of a representational vehicle (e.g. a vector, or a direction in model activation space) determine what it represents? Is content a function of a vector's relationship to environmental inputs, behavioural outputs, training examples, and/or to other representational vehicles? Here, the philosophical literature can be a rich resource for MI practitioners keen to improve their methods for representational content ascription, i.e. precisely and accurately identifying which contents are represented by which aspects of model internals. For example, \citet{Harding2023} operationalises key criteria from philosophical theories of content, in order to guide MI researchers in testing hypotheses about model-represented contents. Her paper makes concrete recommendations for selecting probes, choosing appropriate causal interventions, and inferring representational contents from these methods. For instance, she argues that the method of Iterative Null Space Projection \citep{ravfogel2020null} fails to distinguish between hypothesised and confounding representational content assignments. Philosophical theories can also suggest alternative methods for identifying representational contents, which depart from the information-based approaches discussed by Harding. For example, an alternative family of theories appeals to structural correspondences or morphisms between internal activity and external domains \citep{cummins1996representations, o2004notes, shea2014vi}, and philosophers have explored the application of such theories to identify representational contents in LLMs \citep{Søgaard2023,williams2026structural} and earlier neural networks \citep{o2006connectionist, churchland1998conceptual}.

\paragraph{Suggesting new lines of investigation}
Finally, philosophical research on representation can suggest new avenues for MI research. As \citet{chalmers2025propositional} points out, generic talk of ``features'' tends to obscure the fact that representations in neural networks could in principle represent many different kinds of content: They may represent objects, properties, or relations (concepts, loosely speaking). But they could also represent entire facts or propositions (e.g., that \textit{Paris is the capital of France}). This suggests an open question for MI research --- (how) do models construct propositional representations out of sub-propositional representations? Philosophers also distinguish between a proposition and the attitude of a system towards it—\textit{believing} the proposition \textit{the house is on fire} is quite different from \textit{intending} to bring that proposition about. Whether and how such distinctions are realised in the mechanisms of advanced AI systems are important, but under-investigated questions.% \citep{chalmers2025propositional}.

\subsection{How can we detect deceptive behaviour from model internals?}\label{deception}

One major hope for MI is that it can help to detect unsafe or misaligned model processing that is not evident from outward behaviour \citep{amodei2025urgency}. Here, too, philosophy can make important contributions to MI research. While there are many forms of unsafe or misaligned model processing, we will illustrate our point by focusing on one of the most discussed clusters of issues—deception \citep{hubinger2019risks,park2024ai} and lying \citep{azaria2023internal, pacchiardi2023catch}. MI researchers hope to detect, anticipate, and mitigate deceptive behaviour by AI systems. As \citet{sharkey2025open} argue, ``By monitoring internal representations, [MI methods] could aid in detecting potential sabotage or deceptive behaviour before deployment'' (p. 26), with methods such as linear probing already being investigated \citep{goldowsky-dill2025detecting}. However, the very concept of AI deception raises significant philosophical puzzles \citep{Dung2025}.

\paragraph{Clarifying deception, lying, and related concepts}

While the concepts of deception and lying are easy enough to grasp intuitively, characterising them precisely turns out to be a challenge. But precise, operationalisable and actionable definitions are needed for MI researchers to target the right phenomena. There is a rich literature in ethics and philosophy of language that attempts to clarify the notions of lying and deception, and distinguish them from neighbouring concepts. %Here we give a brief overview of some key insights from this field and how they bear upon attempts to detect lying and deception in AI models using MI.
A standard view of \textit{deception} in philosophy is that it is the act of intentionally causing another agent to form a false belief \citep{sep-lying-definition}. There are two key elements to this definition: (i) the inducement of false belief in another, and (ii) the presence of an intention or goal on the part of the deceiver \citep{carson2010lying, martin2009philosophy}. \textit{Lying} is a related, but distinct concept. Lying is typically taken to involve making a false statement or assertion to someone, where the speaker does not believe the stated claim to be true \citep{sep-lying-definition}. To lie thus requires the speaker makes a specific assertoric commitment to a false proposition, distinguishing it from other speech acts like jokes or questions \citep{marsili2021lying}. \textit{Deception} doesn't always involve lying—one can deceive with one's actions (e.g., ``dummies'' and ``feints'' in sports) or by omitting certain information in conversation without uttering a falsehood. 

One payoff of this philosophical literature for MI is to highlight various ways in which one can make false statements or induce false beliefs without lying or deceiving. For instance, in the cases of jokes, metaphorical statements, fiction-writing, and role play, or mere error, one can utter a falsehood that is not a lie, and in the case of accidentally misleading, one can induce a false belief without deceiving.

Moreover, the literature suggests that lying and deception, as traditionally understood, require significant cognitive complexity. More specifically, they require \textit{intentions} on the part of the deceiver and (in the case of lying) the possession of \textit{beliefs}, and the capacity to make \textit{assertions}. These criteria raise an immediate difficulty when applied to AI systems, as it is highly controversial whether even frontier models possess beliefs, goals, and intentions—or even the ability to make statements or assertions—in the relevant sense. Recent philosophical engagement with these these latter questions thus offers yet another resource to MI researchers concerned with deception and lying in AI models.\footnote{See e.g. \citep{butlin2025ai, williams2024chatting, 10.1093/analys/anag022} on assertion and \citep{WilliamsManuscript-WILIRI-4} on intentions. The question of beliefs in AI models is discussed further in the text below.}

One tempting move is to weaken the definition of lying or deception, to allow that AI systems could lie or deceive without possessing the psychological states or communicative capacities demanded by traditional accounts. For example, if a model persistently outputs misleading answers in contexts where users predictably misinterpret them, we might label that as deception, regardless of the model’s internal ``motivations'' \citep{tarsney2025deception}. However, the idea that insights from MI can help to develop methods for detecting deceptive behaviour based on model internals is premised on a richer notion of lying and deception, one on which the internal states of a system, and not just its behaviour or its effect on users, are relevant. Another approach is to attempt to identify states like beliefs, intentions, and speech acts—or close functional analogues of them—in AI systems. In the next section, we discuss such attempts in MI research and show how philosophy can play a guiding role. (Philosophers have made important contributions to other questions about mental capacities in LLMs. See Box \ref{box}.)

\begin{casebox}[float=t, unbreakable, label=box]{Case Study – Introspection in LLMs}

An illustrative case of productive exchange between philosophers and mechanistic interpretability researchers can be found in recent discussion of \textit{introspection} in language models:

\begin{enumerate}
  \item Early empirical investigations of ``self-knowledge'' and ``introspection'' in LLMs \citep{kadavath2022language,lin2022teaching} measured only the \textit{accuracy} of self-reports. These studies did not distinguish whether models have \textit{direct/privileged access} to internal states or \textit{indirectly infer them}—a central distinction in philosophical discussion of introspection (cf. \cite{schwitzgebel2024introspection}).
  
  \item Philosophers highlighted these conceptual distinctions with a view to informing research into AI/LLM self-knowledge \citep{kammerer2023forms,long2023introspective}. For example, \citet{long2023introspective} flags that models possess substantial extrospective information about systems like themselves, and he proposes evaluation designs which control for this confound.
  
  \item MI researchers absorbed these insights into their subsequent empirical designs, both in collaboration with and with reference to philosophers. Long's \citeyearpar{long2023introspective} proposals led to a collaboration with MI researchers \citep{binder2025looking}, which operationalised privileged access as a model predicting its own behaviour better than a second model trained on its outputs (see also \citep{comsa2025does,song2025privileged}). MI work from Anthropic \citep{lindsey2025emergent} draws on much of this work when operationalising introspection for LLMs.
  
  \item Philosophers made further clarifications relevant to the interpretation of these results. \citet{lederman2026emergent} distinguish \textit{content-agnostic} introspection (e.g. noticing that \textit{some} intervention on one’s inner states has occurred) and \textit{content sensitive} introspection (direct access to the \textit{content} of the injected state) and show that Lindsey’s results establish the former but not the latter.
\end{enumerate}
This example shows a pattern typical of philosophical contributions to MI: philosophers can (i) offer definitions of key constructs, thereby identifying criteria for testing, (ii) suggest finer-grained distinctions to tease apart different empirical targets, with different practical implications (iii) collaborate directly with MI researchers to translate these conceptual insights into experimental designs.

\end{casebox}

\paragraph{Guiding the search for AI “beliefs”}

A key strategy for lie-detection via MI has been to attempt to identify language models' ``beliefs'' from internal states and to detect mismatches between beliefs and outputs. Initial studies used probing classifiers to identify directions in models’ internal activations that correspond to the truth value of inputs. For instance, \citet{azaria2023internal} and \citet{burns2022discovering}, using different probing methods, presented evidence that the truth value of inputs could indeed be decoded from model activations. Azaria and Mitchell frame this project as ``extract[ing] the LLM’s internal belief'' (2023, p. 2) and gave their paper the bold title ``The Internal State of an LLM Knows When It’s Lying''. Should we accept these claims at face value? Belief is a central concept in philosophy, particularly in the sub-fields of philosophy of mind, epistemology, and philosophy of action, so this is a natural place where philosophical engagement can add value to MI research. 

Philosophers Ben Levinstein and Daniel Herrmann \citep{levinstein2024still,herrmann2025standards} have raised questions about the face-value interpretation of the above findings.

For instance, they identified generalisation issues with existing probes, and also argued that for an information-carrying state to qualify as a belief, it must be shown to be used by the system—i.e., causally drive behaviour appropriate to the content of that belief (see also \citealt{Harding2023}). This critique has directly informed subsequent MI research: \citet{marks2023geometry} curated new datasets to address generalisation issues identified by Levinstein and Herrmann and took steps to establish through interventions that these components causally mediate outputs appropriately.

In a follow-up paper, \citet{herrmann2025standards} expand on their previous conceptual work by proposing four criteria for LLM representations to count as beliefs, grounded in existing philosophical literature on belief. They also offer some meta-reflections on the nature and utility of concepts like ``belief'', suggesting that rather than being a binary issue, ``[t]he satisfaction of these requirements come[s] in degrees; in general, the more a representation satisfies these requirements, the more helpful it is to think of the representation as belief-like'' (p. 7). This dialogue illustrates yet another domain in which philosophy can contribute to MI research. By clarifying concepts like \textit{belief}, philosophers can help refine the methods by which MI researchers identify belief-like representations in AI systems, and thus ultimately improve ``lie-detection'' methods.

\paragraph{Drawing ethical and normative boundaries}

MI techniques may be able to identify circuits that are deceptive or suppress information during evaluations. However, not all cases of deception are ethically equivalent. As a result, developers face complex questions about which to intervene on and which to preserve. These decisions require normative frameworks that current MI research lacks. 

Consider three hypothetical ``deceptive'' circuits that may be discovered using MI. The first detects regulatory oversight procedures and triggers the concealment of malicious capabilities (e.g., capacity for producing malware). The second suppresses the home addresses of public figures. The third withholds diagnostic information for certain medical prompts (e.g., ``sharp chest pain and difficulty breathing'') and redirects users to seek out emergency medical care. Standard MI approaches might flag all three mechanisms as instances of the same ``deceptive'' mechanism since they all conceal information contained in the model. Yet, each circuit has distinct ethical implications requiring different interventions. In particular, the first circuit selectively conceals capabilities to circumvent regulatory constraints and should reasonably be targeted for removal. The second circuit protects information that users have no legitimate right to access, and should likely be preserved and potentially strengthened. Finally, the third circuit redirects users to urgent medical care for their own safety and should therefore be preserved and constantly updated to ensure its medical urgency detection is accurate while preserving the core redirection function.

Integrating ethics into MI methodology assist developers in distinguishing cases of information suppression and deception that are ethically problematic and demand intervention from those that are ethically benign \citep[see][]{danaher2022tragic,saetra2021social,kneer2021can,Dung2025}. Our point here isn't that ethicists have superior judgment about which deceptions are justified. It's that normative ethics supplies conceptual frameworks that can help articulate and defend discriminations between types of deception, and identify their ethically relevant dimensions of variation. These dimensions might include (but are not limited to): the \textit{degree of setback} to the interests of the deceived, whether the deceived has a \textit{right to know} the relevant information, the existence of \textit{paternalistic justifications} for deception, and the \textit{established trust/communicative norms} between deceiver and deceived.\footnote{Also relevant here is philosophical work on manipulation, nudging, and epistemic agency \citep{barnhill2022philosophy,pepp2022manipulative}.} MI researchers may already make some of these distinctions implicitly, but where MI's own methodological literature lacks this vocabulary, appeals to those frameworks can make the reasoning public and contestable.

%%%%%%%%%%%%%%%%%%%%%%%%%%%%%%%%%%%%%%%%%%%%%%%%%%%%%%%%%%%%%%%%%%%%%%%
%%%%%%%%%%%%%%%%%%%%%%%%%%%%%%%%%%%%%%%%%%%%%%%%%%%%%%%%%%%%%%%%%%%%%%%
%%%%%%%%%%%%%%%%%%%%%%%%%%%%%%%%%%%%%%%%%%%%%%%%%%%%%%%%%%%%%%%%%%%%%%%
%%%%%%%%%%%%%%%%%%%%%%%%%%%%%%%%%%%%%%%%%%%%%%%%%%%%%%%%%%%%%%%%%%%%%%%
%%%%%%%%%%%%%%%%%%%%%%%%%%%%%%%%%%%%%%%%%%%%%%%%%%%%%%%%%%%%%%%%%%%%%%%

\section{Objections and responses}\label{sec:objections} 

The proposal that philosophy has a central role to play in mechanistic interpretability (MI) may be met with skepticism. We address six common objections below and argue why they do not undermine the value of sustained philosophical engagement in the field.

\paragraph{“Armchair theorising won’t get us anywhere.”}
A common worry is that philosophical contributions are overly abstract or speculative. These ``armchair'' exercises are disconnected from empirical reality and risk missing the complexities of real-world interpretability problems. We think that this is a very valid worry for some philosophical approaches that emphasise \textit{a priori} reasoning and ``in-principle'' arguments. However, much contemporary philosophy is deeply engaged with empirical research.\footnote{For bibliometric evidence, see \citep{prinzing2026role}.} Many of the philosophical works cited in this paper are informed by close engagement with bleeding-edge empirical findings. Some philosophers even get their hands dirty attempting replications of MI studies \citep{levinstein2024still, lederman2026emergent} and operationalising concepts for empirical investigation \citep{Harding2023}. Armchair philosophy of mechanisitic interpretability may be a non-starter, but philosophy need not be practised from the armchair.

\paragraph{“MI researchers can do the philosophising themselves.”}
A second potential objection is that the philosophical dimensions of MI can just as well be addressed internally by MI researchers without needing philosophers. We grant that scientists can contribute to answering philosophical questions about their field, especially those related to the foundations, methods, and implications of scientific practice.\footnote{\citep{turing1950} is a notable early example within the field of computer science.} But while many MI researchers do engage with philosophical questions (as exemplified by \citet{sharkey2025open}), doing so effectively requires philosophical training and knowledge of philosophical literature. As illustrated in this paper, philosophy has many tools, distinctions, and debates that are unfamiliar or underutilised in ML.

And crucially, because philosophers are not burdened with technical implementation, they are often better placed to ``see the forest for the trees'', question framing assumptions, and ask big-picture questions \citep{bickle2025might}. Collaboration between MI researchers and philosophers make the most of each party's expertise, and can mitigate the ever-present risk of reinventing the wheel.

\paragraph{“Philosophers aren’t well informed about MI.”}
To be clear, our claim that mechanistic interpretability needs philosophy should not be read as an injunction aimed solely at MI researchers. Realising the potential contribution of philosophy in MI will equally require more philosophers to direct their efforts and attention to MI, and develop the technical literacy to do so usefully. That said, while many philosophers currently lack the relevant expertise, there is a growing cohort with interdisciplinary training, including backgrounds in computer science, neuroscience, or cognitive modelling. Many philosophy of science programs require coursework in both philosophy and specific scientific disciplines, ensuring that students gain technical foundations alongside philosophical training. And communities of empirically-oriented, technically-literate philosophers exist within most branches of philosophy, including philosophy of mind, epistemology, ethics and philosophy of language.\footnote{This paper has discussed many examples of technically-literate philosophical work that engages with MI research. For additional examples see \citep{beckmann2025new,beckmann2026mechanistic,budding2025large,mollo2026vectorgroundingproblem,goldstein2024doeschatgptmind,grindrod2026sparse,keeling2025attribution,milliere2024philosophical,YetmanForthcoming-YETRIL-2}.} Moreover, as noted above, the key to productive collaboration is not perfect symmetry of expertise, but mutual recognition of complementary strengths.

\paragraph{``Philosophy can foster progress in MI? Prove it!''}
The claim that philosophy has the potential to catalyse progress in
mechanistic interpretability may be met with suspicion, absent a proven track record. But a number of prominent papers in the field have engaged profitably with philosophical work. For instance, influential MI work on causal abstraction and causal mediation analysis, such as \citep{geiger2022inducing, geiger2024finding, geiger2025causal} and \citep{mueller2024missed, mueller2024quest}, cites philosophical work on causation and explanation \citep{hume1748enquiry, pearl2000causality, pearl2001direct, pearl2009causality, salmon1984scientific, lewis1973causation, lewis2000causation, woodward2005making, woodward2021explanatory, halpern2016actual, halpern2016sufficient, yablo1992mental, creel2020understanding}. Recent discussion of ``introspection'' in LLMs provides another example (see Box \ref{box}).

\paragraph{“MI researchers already engage with philosophy.”}
One might turn the previous objection on its head: MI already cites philosophy, so there's no problem to be discussed here. However, the handful of MI papers just discussed represent exceptions, not the rule. While philosophers are increasingly engaging with MI research, MI engagement with philosophy remains a minority trend. As we've illustrated in this paper, there are countless avenues for philosophical contributions to MI that remain unexplored, and philosophical insights have yet to be absorbed into the field at large. Thus, we contend that philosophy's full impact on the field of MI has yet to be realised.

\paragraph{``What is the practical suggestion here?''}
The claim that MI needs philosophy might seem intractably vague. We close with some concrete recommendations, in increasing order of institutional ambition. \textit{At the level of individual practice}, MI researchers can read more philosophy;\footnote{For authoritative and up-to-date overviews of philosophical topic areas, we recommend the Stanford Encyclopedia of Philosophy; the journal \textit{Philosophy Compass}; \textit{Oxford University Press} and \textit{Routledge}'s ``Handbook'', ``Companion'' and ``Contemporary Introduction'' series in philosophy; and \textit{Cambridge University Press} ``Elements'' series. For philosophers actively working on the interpretation of MI results and methods, see many of authors cited in this paper.} attend philosophy workshops and seminars; convene interdisciplinary reading groups; and seek philosophical feedback on manuscript drafts. Philosophers, reciprocally, can engage seriously with MI papers (and develop the technical literacy to do so); follow the MI literature; and write for venues that MI researchers read. \textit{At the level of research groups and institutions}, we recommend co-supervision of graduate students across departments; embedding philosophers-in-residence at MI labs (and reciprocal visits by MI researchers to philosophy departments); summer schools and fellowships designed explicitly for cross-training; co-authored empirical-conceptual projects; and grant calls that require interdisciplinary teams.
And \textit{at the level of the field}, we encourage dedicated philosophy workshops and symposia at machine learning conferences with reciprocal MI sessions at philosophy meetings; position-paper tracks in machine learning venues (like the one hosting this paper), with editorial and reviewer expertise in philosophy; interdisciplinary journals; and undergraduate and graduate curricula that train researchers in both technical and philosophical methods.

We recognise that MI researchers vary in their prior familiarity with philosophy, access to philosophers, and available time and resources to explore research beyond their primary field. However many steps should be attainable for even the busiest MI researchers with little prior engagement with philosophy. For instance, while reading primary literature in philosophy might be infeasible or unfruitful for some MI researchers, they might find it easier to read and apply the insights of more applied philosophical papers that do much of this bridging work for them. And while not all MI researchers will have the time or interest to attend dedicated philosophy conferences, they may find it much easier to attend philosophy tracks and workshops at popular machine learning conferences.\footnote{See e.g. the PhilML workshop this year @ ICML2026.} Finally, while we have imagined ourselves speaking mostly to MI researchers in this paper, we want to emphasise that we view the productive integration of philosophy and mechanistic interpretability as a two-way street: while there are steps MI researchers can take to engage with philosophy, it is equally incumbent on philosophers to make their work useful, comprehensible and discoverable for MI researchers.

%%%%%%%%%%%%%%%%%%%%%%%%%%%%%%%%%%%%%%%%%%%%%%%%%%%%%%%%%%%%%%%%%%%%%%%
%%%%%%%%%%%%%%%%%%%%%%%%%%%%%%%%%%%%%%%%%%%%%%%%%%%%%%%%%%%%%%%%%%%%%%%
%%%%%%%%%%%%%%%%%%%%%%%%%%%%%%%%%%%%%%%%%%%%%%%%%%%%%%%%%%%%%%%%%%%%%%%
%%%%%%%%%%%%%%%%%%%%%%%%%%%%%%%%%%%%%%%%%%%%%%%%%%%%%%%%%%%%%%%%%%%%%%%
%%%%%%%%%%%%%%%%%%%%%%%%%%%%%%%%%%%%%%%%%%%%%%%%%%%%%%%%%%%%%%%%%%%%%%%

\section{Conclusion}\label{sec:conclusion}

Mechanistic interpretability is a rapidly evolving field, driven by urgent practical needs and rich with conceptual complexity. As we have argued, philosophy is deeply relevant to this field. MI raises foundational questions about explanation, representation, knowledge, agency, and values, and meaningful progress in the field is often predicated on handling these concepts carefully. As illustrated above, philosophers can help provide conceptual clarity, identifying and scrutinizing assumptions, proposing novel research questions, interpreting empirical results, and illuminating ethical complexities. To stress, our claim that mechanistic interpretability needs philosophy is neither a call for disciplinary silos to remain intact, nor a claim that philosophy has ready-made answers for all of the challenges encountered in MI research. Instead, it is an invitation to deeper interdisciplinary collaboration that is technically informed and philosophically grounded. As AI systems become more powerful and more deeply embedded in society, the stakes of understanding them, not just how they behave, but how they work, have never been higher. Enriched by philosophy, mechanistic interpretability has a clearer shot of success.

\begin{ack}
We are grateful to Thor Grünbaum, Andy Mckilliam, participants in the technology and philosophy reading group at the University of Copenhagen, and an audience at the 6th Conference on Philosophy of Artificial Intelligence in Amsterdam, for feedback and discussion. We also thank three anonymous referees and area chair at NeurIPS, for their valuable comments.

\textbf{Funding.} This work was supported by funding from the Carlsberg Foundation (Grant CF22-1432). CF was supported by Danish Data Science Academy, which is funded by the Novo Nordisk Foundation (NNF21SA0069429).

\textbf{Competing interests.} The authors declare no competing interests.
\end{ack}

\bibliography{refs_clean}

@inproceedings{balkir-etal-2022-necessity,
    title = "Necessity and Sufficiency for Explaining Text Classifiers: A Case Study in Hate Speech Detection",
    author = "Balkir, Esma  and
      Nejadgholi, Isar  and
      Fraser, Kathleen  and
      Kiritchenko, Svetlana",
    editor = "Carpuat, Marine  and
      de Marneffe, Marie-Catherine  and
      Meza Ruiz, Ivan Vladimir",
    booktitle = "Proceedings of the 2022 Conference of the North American Chapter of the Association for Computational Linguistics: Human Language Technologies",
    month = jul,
    year = "2022",
    address = "Seattle, United States",
    publisher = "Association for Computational Linguistics",
    url = "https://aclanthology.org/2022.naacl-main.192/",
    doi = "10.18653/v1/2022.naacl-main.192",
    pages = "2672--2686"
}

@incollection{barnhill2022philosophy,
  title={How philosophy might contribute to the practical ethics of online manipulation},
  author={Barnhill, Anne},
  booktitle={The philosophy of online manipulation},
  editor={Jongepier, Fleur and Klenk, Michael},
  pages={49--71},
  year={2022},
  publisher={Routledge}
}

@article{mitchell2023landscape,
  title={The landscape of integrative pluralism},
  author={Mitchell, Sandra D},
  journal={Theoria: An International Journal for Theory, History and Foundations of Science},
  volume={38},
  number={3},
  pages={261--297},
  year={2023},
  publisher={JSTOR}
}

@inproceedings{budding2024does,
  title={Does Explainable {AI} Need Cognitive Models?},
  author={Budding, C{\'e}line and Zednik, Carlos},
  booktitle={Proceedings of the Annual Meeting of the Cognitive Science Society},
  volume={46},
  year={2024}
}

@book{darden2006reasoning,
  title={Reasoning in biological discoveries: Essays on mechanisms, interfield relations, and anomaly resolution},
  author={Darden, Lindley},
  year={2006},
  publisher={Cambridge University Press}
}

@book{carson2010lying,
  title={Lying and deception: Theory and practice},
  author={Carson, Thomas L},
  year={2010},
  publisher={Oxford University Press}
}

@incollection{pepp2022manipulative,
  title={Manipulative machines},
  author={Pepp, Jessica and Sterken, Rachel and McKeever, Matthew and Michaelson, Eliot},
  booktitle={The philosophy of online manipulation},
  pages={91--107},
  year={2022},
  publisher={Routledge}
}

@book{martin2009philosophy,
  title={The philosophy of deception},
  author={Martin, Clancy W},
  year={2009},
  publisher={Oxford University Press}
}

@article{park2024ai,
  title={{AI} deception: A survey of examples, risks, and potential solutions},
  author={Park, Peter S and Goldstein, Simon and O’Gara, Aidan and Chen, Michael and Hendrycks, Dan},
  journal={Patterns},
  volume={5},
  number={5},
  year={2024},
  publisher={Elsevier}
}

@article{tarsney2025deception,
  title={Deception and manipulation in generative {AI}},
  author={Tarsney, Christian},
  journal={Philosophical Studies},
  pages={1--23},
  year={2025},
  publisher={Springer}
}

@article{turing1950,
  title={Computing Machinery and Intelligence},
  author={Turing, Alan M.},
  journal={Mind},
  volume={59},
  number={236},
  pages={433--460},
  year={1950}
}

@article{chalmers2025propositional,
  title={Propositional interpretability in artificial intelligence},
  author={Chalmers, David J},
  journal={arXiv preprint arXiv:2501.15740},
  year={2025}
}

@book{sogaard2021explainable,
  title={Explainable natural language processing},
  author={S{\o}gaard, Anders},
  year={2021},
  publisher={Morgan \& Claypool Publishers}
}

@article{kotonya2020explainable,
  title={Explainable automated fact-checking: A survey},
  author={Kotonya, Neema and Toni, Francesca},
  journal={arXiv preprint arXiv:2011.03870},
  year={2020}
}

@book{molnar2020interpretable,
  title={Interpretable machine learning},
  author={Molnar, Christoph},
  year={2020},
  publisher={Lulu.com}
}

@article{carvalho2019machine,
  title={Machine learning interpretability: A survey on methods and metrics},
  author={Carvalho, Diogo V and Pereira, Eduardo M and Cardoso, Jaime S},
  journal={Electronics},
  volume={8},
  number={8},
  pages={832},
  year={2019},
  publisher={Multidisciplinary Digital Publishing Institute}
}

@article{guidotti2018survey,
  title={A survey of methods for explaining black box models},
  author={Guidotti, Riccardo and Monreale, Anna and Ruggieri, Salvatore and Turini, Franco and Giannotti, Fosca and Pedreschi, Dino},
  journal={ACM computing surveys (CSUR)},
  volume={51},
  number={5},
  pages={1--42},
  year={2018},
  publisher={ACM New York, NY, USA}
}

@inproceedings{goebel2018explainable,
  title={Explainable {AI}: the new 42?},
  author={Goebel, Randy and Chander, Ajay and Holzinger, Katharina and Lecue, Freddy and Akata, Zeynep and Stumpf, Simone and Kieseberg, Peter and Holzinger, Andreas},
  booktitle={International cross-domain conference for machine learning and knowledge extraction},
  pages={295--303},
  year={2018},
  organization={Springer}
}

@article{miller2019explanation,
  title={Explanation in artificial intelligence: Insights from the social sciences},
  author={Miller, Tim},
  journal={Artificial intelligence},
  volume={267},
  pages={1--38},
  year={2019},
  publisher={Elsevier}
}

@inproceedings{xu2019explainable,
  title={Explainable {AI}: A brief survey on history, research areas, approaches and challenges},
  author={Xu, Feiyu and Uszkoreit, Hans and Du, Yangzhou and Fan, Wei and Zhao, Dongyan and Zhu, Jun},
  booktitle={Natural language processing and Chinese computing: 8th cCF international conference, NLPCC 2019, dunhuang, China, October 9--14, 2019, proceedings, part II 8},
  pages={563--574},
  year={2019},
  organization={Springer}
}

@article{dwivedi2023explainable,
  title={Explainable {AI} ({XAI}): Core ideas, techniques, and solutions},
  author={Dwivedi, Rudresh and Dave, Devam and Naik, Het and Singhal, Smiti and Omer, Rana and Patel, Pankesh and Qian, Bin and Wen, Zhenyu and Shah, Tejal and Morgan, Graham and others},
  journal={ACM Computing Surveys},
  volume={55},
  number={9},
  pages={1--33},
  year={2023},
  publisher={ACM New York, NY}
}

@article{ghosh2020interpretable,
  title={Interpretable artificial intelligence: why and when},
  author={Ghosh, Adarsh and Kandasamy, Devasenathipathy},
  journal={American Journal of Roentgenology},
  volume={214},
  number={5},
  pages={1137--1138},
  year={2020},
  publisher={American Roentgen Ray Society}
}

@article{sathyan2022interpretable,
  title={Interpretable {AI} for bio-medical applications},
  author={Sathyan, Anoop and Weinberg, Abraham Itzhak and Cohen, Kelly},
  journal={Complex engineering systems (Alhambra, Calif.)},
  volume={2},
  number={4},
  pages={18},
  year={2022}
}

@article{shin2022platforms,
  title={In platforms we trust? {U}nlocking the black-box of news algorithms through interpretable {AI}},
  author={Shin, Donghee and Zaid, Bouziane and Biocca, Frank and Rasul, Azmat},
  journal={Journal of Broadcasting \& Electronic Media},
  volume={66},
  number={2},
  pages={235--256},
  year={2022},
  publisher={Taylor \& Francis}
}

@inproceedings{rauker2023toward,
  title={Toward transparent {AI}: A survey on interpreting the inner structures of deep neural networks},
  author={R{\"a}uker, Tilman and Ho, Anson and Casper, Stephen and Hadfield-Menell, Dylan},
  booktitle={2023 ieee conference on secure and trustworthy machine learning (satml)},
  pages={464--483},
  year={2023},
  organization={IEEE}
}

@inproceedings{10.5555/3692070.3694093,
author = {Vilas, Martina G. and Adolfi, Federico and Poeppel, David and Roig, Gemma},
title = {Position: an inner interpretability framework for {AI} inspired by lessons from cognitive neuroscience},
year = {2024},
publisher = {JMLR.org},
booktitle = {Proceedings of the 41st International Conference on Machine Learning},
articleno = {2023},
numpages = {17},
location = {Vienna, Austria},
series = {ICML'24}
}

@article{grzankowski2024real,
  title={Real sparks of artificial intelligence and the importance of inner interpretability},
  author={Grzankowski, Alex},
  journal={Inquiry},
  pages={1--27},
  year={2024},
  publisher={Taylor \& Francis}
}

@inproceedings{lakkaraju2019faithful,
  title={Faithful and customizable explanations of black box models},
  author={Lakkaraju, Himabindu and Kamar, Ece and Caruana, Rich and Leskovec, Jure},
  booktitle={Proceedings of the 2019 AAAI/ACM Conference on AI, Ethics, and Society},
  pages={131--138},
  year={2019}
}

@article{calderon2024behalf,
  title={On behalf of the stakeholders: Trends in {NLP} model interpretability in the era of {LLMs}},
  author={Calderon, Nitay and Reichart, Roi},
  journal={arXiv preprint arXiv:2407.19200},
  year={2024}
}

@inproceedings{
meloux2025everything,
title={Everything, Everywhere, All at Once: Is Mechanistic Interpretability Identifiable?},
author={Maxime M{\'e}loux and Silviu Maniu and Fran{\c{c}}ois Portet and Maxime Peyrard},
booktitle={The Thirteenth International Conference on Learning Representations},
year={2025},
url={https://openreview.net/forum?id=5IWJBStfU7}
}

@inproceedings{adolfi2024complexity,
  title={Complexity-theoretic limits on the promises of artificial neural network reverse-engineering},
  author={Adolfi, Federico and Vilas, Martina G and Wareham, Todd},
  booktitle={Proceedings of the Annual Meeting of the Cognitive Science Society},
  volume={46},
  year={2024}
}

@article{laplane2019science,
  title={Why science needs philosophy},
  author={Laplane, Lucie and Mantovani, Paolo and Adolphs, Ralph and Chang, Hasok and Mantovani, Alberto and McFall-Ngai, Margaret and Rovelli, Carlo and Sober, Elliott and Pradeu, Thomas},
  journal={Proceedings of the National Academy of Sciences},
  volume={116},
  number={10},
  pages={3948--3952},
  year={2019},
  publisher={National Academy of Sciences}
}

@inproceedings{
makelov2024is,
title={Is This the Subspace You Are Looking for? An Interpretability Illusion for Subspace Activation Patching},
author={Aleksandar Makelov and Georg Lange and Atticus Geiger and Neel Nanda},
booktitle={The Twelfth International Conference on Learning Representations},
year={2024},
url={https://openreview.net/forum?id=Ebt7JgMHv1}
}

@article{fierro2024defining,
  title={Defining knowledge: Bridging epistemology and large language models},
  author={Fierro, Constanza and Dhar, Ruchira and Stamatiou, Filippos and Garneau, Nicolas and S{\o}gaard, Anders},
  journal={Proceedings of the 2024 conference on empirical methods in natural language processing},
  year={2024}
}

@article{rudin2019stop,
  title={Stop explaining black box machine learning models for high stakes decisions and use interpretable models instead},
  author={Rudin, Cynthia},
  journal={Nature machine intelligence},
  volume={1},
  number={5},
  pages={206--215},
  year={2019},
  publisher={Nature Publishing Group UK London}
}

@article{sharkey2025open,
  title={Open Problems in Mechanistic Interpretability},
  author={Sharkey, Lee and Chughtai, Bilal and Batson, Joshua and Lindsey, Jack and Wu, Jeff and Bushnaq, Lucius and Goldowsky-Dill, Nicholas and Heimersheim, Stefan and Ortega, Alejandro and Bloom, Joseph and others},
  journal={arXiv preprint arXiv:2501.16496},
  year={2025}
}

@article{elhage2022toy,
  title={Toy models of superposition},
  author={Elhage, Nelson and Hume, Tristan and Olsson, Catherine and Schiefer, Nicholas and Henighan, Tom and Kravec, Shauna and Hatfield-Dodds, Zac and Lasenby, Robert and Drain, Dawn and Chen, Carol and others},
  journal={arXiv preprint arXiv:2209.10652},
  year={2022}
}

@article{geiger2025causal,
  title={Causal abstraction: A theoretical foundation for mechanistic interpretability},
  author={Geiger, Atticus and Ibeling, Duligur and Zur, Amir and Chaudhary, Maheep and Chauhan, Sonakshi and Huang, Jing and Arora, Aryaman and Wu, Zhengxuan and Goodman, Noah and Potts, Christopher and others},
  journal={Journal of Machine Learning Research},
  volume={26},
  year={2025}
}

@article{kastner2024explaining,
  title={Explaining {AI} through mechanistic interpretability},
  author={K{\"a}stner, Lena and Crook, Barnaby},
  journal={European Journal for Philosophy of Science},
  volume={14},
  number={4},
  pages={52},
  year={2024},
  publisher={Springer}
}

@article{saphra2024mechanistic,
  title={Mechanistic?},
  author={Saphra, Naomi and Wiegreffe, Sarah},
  journal={Proceedings of the 7th BlackboxNLP Workshop: Analyzing and Interpreting Neural Networks for NLP},
  year={2024}
}

@article{bereska2024mechanistic,
  title={Mechanistic Interpretability for {AI} Safety--A Review},
  author={Bereska, Leonard and Gavves, Efstratios},
  journal={Transactions in Machine Learning Research},
  year={2024}
}

@article{olsson2022context,
  title={In-context learning and induction heads},
  author={Olsson, Catherine and Elhage, Nelson and Nanda, Neel and Joseph, Nicholas and DasSarma, Nova and Henighan, Tom and Mann, Ben and Askell, Amanda and Bai, Yuntao and Chen, Anna and others},
  journal={arXiv preprint arXiv:2209.11895},
  year={2022}
}

@article{machamer2000thinking,
  title={Thinking about mechanisms},
  author={Machamer, Peter and Darden, Lindley and Craver, Carl F},
  journal={Philosophy of science},
  volume={67},
  number={1},
  pages={1--25},
  year={2000},
  publisher={Cambridge University Press}
}

@article{bechtel2005explanation,
  title={Explanation: A mechanist alternative},
  author={Bechtel, William and Abrahamsen, Adele},
  journal={Studies in History and Philosophy of Science Part C: Studies in History and Philosophy of Biological and Biomedical Sciences},
  volume={36},
  number={2},
  pages={421--441},
  year={2005},
  publisher={Elsevier}
}

@book{craver2007explaining,
  title={Explaining the brain: Mechanisms and the mosaic unity of neuroscience},
  author={Craver, Carl F},
  year={2007},
  publisher={Clarendon Press}
}

@book{woodward2005making,
  title={Making things happen: A theory of causal explanation},
  author={Woodward, James},
  year={2005},
  publisher={Oxford university press}
}

@book{craver2013search,
  title={In search of mechanisms: Discoveries across the life sciences},
  author={Craver, Carl F and Darden, Lindley},
  year={2013},
  publisher={University of Chicago Press}
}

@inproceedings{azaria2023internal,
  title={The Internal State of an {LLM} Knows When It’s Lying},
  author={Azaria, Amos and Mitchell, Tom},
  booktitle={Findings of the Association for Computational Linguistics: EMNLP 2023},
  pages={967--976},
  year={2023}
}

@article{burns2022discovering,
  title={Discovering latent knowledge in language models without supervision},
  author={Burns, Collin and Ye, Haotian and Klein, Dan and Steinhardt, Jacob},
  journal={The Eleventh International Conference on Learning Representations},
  year={2023}
}

@article{levinstein2024still,
  title={Still no lie detector for language models: Probing empirical and conceptual roadblocks},
  author={Levinstein, Benjamin A and Herrmann, Daniel A},
  journal={Philosophical Studies},
  pages={1--27},
  year={2024},
  publisher={Springer}
}

@article{herrmann2025standards,
  title={Standards for belief representations in {LLMs}},
  author={Herrmann, Daniel A and Levinstein, Benjamin A},
  journal={Minds and Machines},
  volume={35},
  number={1},
  pages={1--25},
  year={2025},
  publisher={Springer}
}

@article{Harding2023,
author = {Harding, Jacqueline},
doi = {10.1086/728685},
journal = {The British Journal for the Philosophy of Science},
pages = {1--36},
title = {{Operationalising Representation in Natural Language Processing}},
year = {2023}
}

@inproceedings{marks2023geometry,
  title={The Geometry of Truth: Emergent Linear Structure in Large Language Model Representations of True/False Datasets},
  author={Marks, Samuel and Tegmark, Max},
  booktitle={First Conference on Language Modeling},
  year={2024}
}

@article{keeling2025attribution,
  title={On the attribution of confidence to large language models},
  author={Keeling, Geoff and Street, Winnie},
  journal={Inquiry},
  pages={1--27},
  year={2025},
  publisher={Taylor \& Francis}
}

@article{mueller2024quest,
    author = {Mueller, Aaron and Brinkmann, Jannik and Li, Millicent and Marks, Samuel and Pal, Koyena and Prakash, Nikhil and Rager, Can and Sankaranarayanan, Aruna and Sharma, Arnab Sen and Sun, Jiuding and Todd, Eric and Bau, David and Belinkov, Yonatan},
    title = {The Quest for the Right Mediator: Surveying Mechanistic Interpretability for NLP Through the Lens of Causal Mediation Analysis},
    journal = {Computational Linguistics},
    volume = {52},
    number = {1},
    pages = {331-378},
    year = {2026},
    month = {03},
    issn = {0891-2017},
    doi = {10.1162/COLI.a.572},
    url = {https://doi.org/10.1162/COLI.a.572},
    eprint = {https://direct.mit.edu/coli/article-pdf/52/1/331/2554934/coli.a.572.pdf},
}

@article{olah2020zoom,
  title={Zoom in: An introduction to circuits},
  author={Olah, Chris and Cammarata, Nick and Schubert, Ludwig and Goh, Gabriel and Petrov, Michael and Carter, Shan},
  journal={Distill},
  volume={5},
  number={3},
  pages={e00024--001},
  year={2020}
}

@incollection{craver2014levels,
  title={Levels},
  author={Craver, Carl F},
  booktitle={Open {MIND}},
  year={2015},
  publisher={Frankfurt am Main: MIND Group}
}

@book{shea2018representation,
  title={Representation in cognitive science},
  author={Shea, Nicholas},
  year={2018},
  publisher={Oxford University Press}
}

@article{danaher2022tragic,
  title={Tragic choices and the virtue of techno-responsibility gaps},
  author={Danaher, John},
  journal={Philosophy \& Technology},
  volume={35},
  number={2},
  pages={26},
  year={2022},
  publisher={Springer}
}

@inproceedings{zeiler2014visualizing,
  title={Visualizing and understanding convolutional networks},
  author={Zeiler, Matthew D and Fergus, Rob},
  booktitle={Computer Vision--ECCV 2014: 13th European Conference, Zurich, Switzerland, September 6-12, 2014, Proceedings, Part I 13},
  pages={818--833},
  year={2014},
  organization={Springer}
}

@article{nezhurina2024alice,
  title={Alice in wonderland: Simple tasks showing complete reasoning breakdown in state-of-the-art large language models},
  author={Nezhurina, Marianna and Cipolina-Kun, Lucia and Cherti, Mehdi and Jitsev, Jenia},
  journal={arXiv preprint arXiv:2406.02061},
  year={2025}
}

@article{lewis2024evaluating,
  title={Evaluating the Robustness of Analogical Reasoning in Large Language Models},
  author={Lewis, Martha and Mitchell, Melanie},
  journal={Transactions on Machine Learning Research},
  year={2025}
}

@article{ullman2023large,
  title={Large language models fail on trivial alterations to theory-of-mind tasks},
  author={Ullman, Tomer},
  journal={arXiv preprint arXiv:2302.08399},
  year={2023}
}

@article{strachan2024testing,
  title={Testing theory of mind in large language models and humans},
  author={Strachan, James WA and Albergo, Dalila and Borghini, Giulia and Pansardi, Oriana and Scaliti, Eugenio and Gupta, Saurabh and Saxena, Krati and Rufo, Alessandro and Panzeri, Stefano and Manzi, Guido and others},
  journal={Nature Human Behaviour},
  volume={8},
  number={7},
  pages={1285--1295},
  year={2024},
  publisher={Nature Publishing Group UK London}
}

@article{geiger2021causal,
  title={Causal abstractions of neural networks},
  author={Geiger, Atticus and Lu, Hanson and Icard, Thomas and Potts, Christopher},
  journal={Advances in Neural Information Processing Systems},
  volume={34},
  pages={9574--9586},
  year={2021}
}

@article{kaplan2011explanatory,
  title={The explanatory force of dynamical and mathematical models in neuroscience: A mechanistic perspective},
  author={Kaplan, David Michael and Craver, Carl F},
  journal={Philosophy of science},
  volume={78},
  number={4},
  pages={601--627},
  year={2011},
  publisher={Cambridge University Press}
}

@article{piccinini2011integrating,
  title={Integrating psychology and neuroscience: Functional analyses as mechanism sketches},
  author={Piccinini, Gualtiero and Craver, Carl},
  journal={Synthese},
  volume={183},
  pages={283--311},
  year={2011},
  publisher={Springer}
}

@misc{Williams_2024, title={“{W}hat the hell is a representation, anyway?” | {C}larifying {AI} interpretability with tools from philosophy of cognitive science | {P}art 1: {V}ehicles vs. contents}, url={https://www.lesswrong.com/posts/h43tWdo79C6dzXf8x/what-the-hell-is-a-representation-anyway-or-clarifying-ai?utm_campaign=post_share&utm_source=link}, journal={LessWrong}, author={Williams, Iwan}, year={2024}, month={Jun}}

@article{shea2007content,
  title={Content and its vehicles in connectionist systems},
  author={Shea, Nicholas},
  journal={Mind \& Language},
  volume={22},
  number={3},
  pages={246--269},
  year={2007},
  publisher={Wiley Online Library}
}

@article{azhar2016polytopes,
  title={Polytopes as vehicles of informational content in feedforward neural networks},
  author={Azhar, Feraz},
  journal={Philosophical Psychology},
  volume={29},
  number={5},
  pages={697--716},
  year={2016},
  publisher={Taylor \& Francis}
}

@book{clark1993associative,
  title={Associative engines: Connectionism, concepts, and representational change},
  author={Clark, Andy},
  year={1993},
  publisher={MIT Press}
}

@book{cummins1996representations,
  title={Representations, targets, and attitudes},
  author={Cummins, Robert},
  year={1996},
  publisher={MIT press}
}

@incollection{o2004notes,
  title={Notes toward a structuralist theory of mental representation},
  author={O'Brien, Gerard and Opie, Jon},
  booktitle={Representation in mind},
  pages={1--20},
  year={2004},
  publisher={Elsevier}
}

@article{o2006connectionist,
  title={How do connectionist networks compute?},
  author={O’Brien, Gerard and Opie, Jon},
  journal={Cognitive Processing},
  volume={7},
  pages={30--41},
  year={2006},
  publisher={Springer}
}

@inproceedings{shea2014vi,
  title={Exploitable isomorphism and structural representation},
  author={Shea, Nicholas},
  booktitle={Proceedings of the Aristotelian Society},
  volume={114},
  number={2\_pt\_2},
  pages={123--144},
  year={2014},
  organization={Oxford University Press Oxford, UK}
}

@article{Søgaard2023,
author = {S{\o}gaard, Anders},
doi = {10.1007/s11023-023-09622-4},
journal = {Minds and Machines},
number = {1},
pages = {33--54},
title = {{Grounding the Vector Space of an Octopus: Word Meaning from Raw Text}},
url = {https://doi.org/10.1007/s11023-023-09622-4},
volume = {33},
year = {2023}
}

@article{churchland1998conceptual,
  title={Conceptual similarity across sensory and neural diversity: The {F}odor/{L}epore challenge answered},
  author={Churchland, Paul M},
  journal={The Journal of Philosophy},
  volume={95},
  number={1},
  pages={5--32},
  year={1998},
  publisher={JSTOR}
}

@article{bickle2025might,
  title={How Might a ‘Philosopher's Toolkit’ {H}elp Advance Neuroscience? {L}et's Ask Some Neuroscientists},
  author={Bickle, John and Churchland, Patricia and Firestein, Stuart and Lehman, Michael N and Parker, David J and Silva, Alcino J and Walters, Bradley J and Williams, Robert W},
  journal={European Journal of Neuroscience},
  volume={61},
  number={7},
  pages={e70074},
  year={2025},
  publisher={Wiley Online Library}
}

@article{rovelli2018physics,
  title={Physics needs philosophy. Philosophy needs physics},
  author={Rovelli, Carlo},
  journal={Foundations of Physics},
  volume={48},
  number={5},
  pages={481--491},
  year={2018},
  publisher={Springer}
}

@article{thagard2009cognitive,
  title={Why cognitive science needs philosophy and vice versa},
  author={Thagard, Paul},
  journal={Topics in Cognitive Science},
  volume={1},
  number={2},
  pages={237--254},
  year={2009},
  publisher={Wiley Online Library}
}

@article{nussbaum2016economics,
  title={Economics still needs philosophy},
  author={Nussbaum, Martha C},
  journal={Review of Social Economy},
  volume={74},
  number={3},
  pages={229--247},
  year={2016},
  publisher={Taylor \& Francis}
}

@book{buckner2024deep,
  title={From deep learning to rational machines: What the history of philosophy can teach us about the future of artificial intelligence},
  author={Buckner, Cameron J},
  year={2024},
  publisher={Oxford University Press}
}

@article{hubinger2019risks,
  title={Risks from learned optimization in advanced machine learning systems},
  author={Hubinger, Evan and van Merwijk, Chris and Mikulik, Vladimir and Skalse, Joar and Garrabrant, Scott},
  journal={arXiv preprint arXiv:1906.01820},
  year={2019}
}

@article{Glennan2022,
  title={Six theses on mechanisms and mechanistic science},
  author={Glennan, Stuart and Illari, Phyllis and Weber, Erik},
  journal={Journal for General Philosophy of Science},
  volume={53},
  number={2},
  pages={143--161},
  year={2022},
  publisher={Springer}
}

@incollection{darden2017strategies,
  title={Strategies for discovering mechanisms},
  author={Darden, Lindley},
  booktitle={The Routledge handbook of mechanisms and mechanical philosophy},
  pages={255--266},
  year={2017},
  publisher={Routledge}
}

@incollection{ryder2009problems1,
  title={Problems of representation I: nature and role},
  author={Ryder, Dan},
  booktitle={The Routledge companion to philosophy of psychology},
  pages={233--250},
  year={2009},
  publisher={Routledge}
}

@incollection{ryder2009problems2,
  title={Problems of representation II: Naturalizing content},
  author={Ryder, Dan},
  booktitle={The Routledge companion to philosophy of psychology},
  pages={251--279},
  year={2009},
  publisher={Routledge}
}

@book{schulte2023mental,
  title={Mental content},
  author={Schulte, Peter},
  year={2023},
  publisher={Cambridge University Press}
}

@InCollection{sep-lying-definition,
	author       =	{Mahon, James Edwin},
	title        =	{{The Definition of Lying and Deception}},
	booktitle    =	{The {Stanford} Encyclopedia of Philosophy},
	editor       =	{Edward N. Zalta},
	howpublished =	{\url{https://plato.stanford.edu/archives/win2016/entries/lying-definition/}},
	year         =	{2016},
	edition      =	{{W}inter 2016},
	publisher    =	{Metaphysics Research Lab, Stanford University}
}

@article{pacchiardi2023catch,
  title={How to catch an {AI} liar: Lie detection in black-box {LLMs} by asking unrelated questions},
  author={Pacchiardi, Lorenzo and Chan, Alex J and Mindermann, S{\"o}ren and Moscovitz, Ilan and Pan, Alexa Y and Gal, Yarin and Evans, Owain and Brauner, Jan},
  journal={The Twelfth International Conference on Learning Representations},
  year={2024}
}

@article{marsili2021lying,
  title={Lying, speech acts, and commitment},
  author={Marsili, Neri},
  journal={Synthese},
  volume={199},
  number={1},
  pages={3245--3269},
  year={2021},
  publisher={Springer}
}

@article{saetra2021social,
  title={Social robot deception and the culture of trust},
  author={S{\ae}tra, Henrik Skaug},
  journal={Paladyn, Journal of Behavioral Robotics},
  volume={12},
  number={1},
  pages={276--286},
  year={2021},
  publisher={De Gruyter}
}

@article{kneer2021can,
  title={Can a robot lie? Exploring the folk concept of lying as applied to artificial agents},
  author={Kneer, Markus},
  journal={Cognitive Science},
  volume={45},
  number={10},
  pages={e13032},
  year={2021},
  publisher={Wiley Online Library}
}

@article{williams2024chatting,
  title={Chatting with bots: {AI}, speech acts, and the edge of assertion},
  author={Williams, Iwan and Bayne, Tim},
  journal={Inquiry},
  pages={1--24},
  year={2024},
  publisher={Taylor \& Francis}
}

@article{Hendrycks_Hiscott_2025, title={The Misguided Quest for Mechanistic {AI} Interpretability}, url={https://www.ai-frontiers.org/articles/the-misguided-quest-for-mechanistic-ai-interpretability}, journal={AI Frontiers}, author={Hendrycks, Dan and Hiscott, Laura}, year={2025}, month={May}}

@article{zhong2023clock,
  title={The clock and the pizza: Two stories in mechanistic explanation of neural networks},
  author={Zhong, Ziqian and Liu, Ziming and Tegmark, Max and Andreas, Jacob},
  journal={Advances in neural information processing systems},
  volume={36},
  pages={27223--27250},
  year={2023}
}

@article{taylor2022signature,
  title={The signature-testing approach to mapping biological and artificial intelligences},
  author={Taylor, Alex H and Bastos, Amalia PM and Brown, Rachael L and Allen, Colin},
  journal={Trends in Cognitive Sciences},
  volume={26},
  number={9},
  pages={738--750},
  year={2022},
  publisher={Elsevier}
}

@misc{amodei2025urgency,
  author = {Amodei, Dario},
  title = {The Urgency of Interpretability},
  year = {2025},
  howpublished = {\url{https://www.darioamodei.com/post/the-urgency-of-interpretability}},
  note = {Accessed: 2025-05-21}
}

@article{williams2026structural,
  title={Can structural correspondences ground real-world representational content in large language models?},
  author={Williams, Iwan},
  journal={Mind \& Language},
  year={2026},
}

@article{park2023linear,
  title={The linear representation hypothesis and the geometry of large language models},
  author={Park, Kiho and Choe, Yo Joong and Veitch, Victor},
  journal={arXiv preprint arXiv:2311.03658},
  year={2023}
}

@article{butlin2025ai,
  title={{AI} assertion},
  author={Butlin, Patrick and Viebahn, Emanuel},
  journal={Ergo an Open Access Journal of Philosophy},
  volume={12},
  year={2025},
  publisher={Michigan Publishing}
}

@article{10.1093/analys/anag022,
    author = {Skiba, Lukas},
    title = {Artificial Speech Acts},
    journal = {Analysis},
    pages = {anag022},
    year = {2026},
    month = {04},
    doi = {10.1093/analys/anag022},
    url = {https://doi.org/10.1093/analys/anag022},
    eprint = {https://academic.oup.com/analysis/advance-article-pdf/doi/10.1093/analys/anag022/67864057/anag022.pdf},
}

@article{Dung2025,
  author  = {Dung, Leonard},
  title   = {A Two-Step, Multidimensional Account of Deception in Language Models},
  journal = {Erkenntnis},
  year    = {2025},
  month   = oct,
  day     = {13},
  issn    = {1572-8420},
  doi     = {10.1007/s10670-025-01017-4},
  url     = {https://doi.org/10.1007/s10670-025-01017-4},
}

@article{WilliamsManuscript-WILIRI-4,
	author = {Iwan Williams},
	title = {Intention-Like Representations in Language Models?},
	year = {2025},
    journal = {PhilPapers preprint},
}

@inproceedings{geiger2022inducing,
  author    = {Geiger, Atticus and Wu, Zhengxuan and Lu, Hanson and
               Rozner, Josh and Kreiss, Elisa and Icard, Thomas and
               Goodman, Noah D. and Potts, Christopher},
  title     = {Inducing Causal Structure for Interpretable Neural Networks},
  booktitle = {Proceedings of the 39th International Conference on
               Machine Learning ({ICML})},
  series    = {Proceedings of Machine Learning Research},
  volume    = {162},
  pages     = {7324--7338},
  year      = {2022},
  publisher = {PMLR},
}

@inproceedings{geiger2024finding,
  author    = {Geiger, Atticus and Wu, Zhengxuan and Potts, Christopher and
               Icard, Thomas and Goodman, Noah D.},
  title     = {Finding Alignments Between Interpretable Causal Variables
               and Distributed Neural Representations},
  booktitle = {Causal Learning and Reasoning ({CLeaR})},
  series    = {Proceedings of Machine Learning Research},
  volume    = {236},
  year      = {2024},
  publisher = {PMLR},
}

@misc{mueller2024missed,
  author    = {Mueller, Aaron},
  title     = {Missed Causes and Ambiguous Effects: Counterfactuals Pose
               Challenges for Interpreting Neural Networks},
  year      = {2024},
  eprint    = {2407.04690},
  archivePrefix = {arXiv},
  note      = {ICML 2024 Mechanistic Interpretability Workshop,
               Honorable Mention},
}

@book{hume1748enquiry,
  author    = {Hume, David},
  title     = {An Enquiry Concerning Human Understanding},
  year      = {1748},
  publisher = {A. Millar},
  address   = {London},
}

@incollection{lewis1973causation,
  author    = {Lewis, David},
  title     = {Causation},
  booktitle = {Philosophical Papers, {Volume II}},
  editor    = {Lewis, David},
  publisher = {Oxford University Press},
  pages     = {159--213},
  year      = {1986},
  note      = {Originally published in \emph{Journal of Philosophy}
               70(17):556--567, 1973},
}

@article{lewis2000causation,
  author    = {Lewis, David},
  title     = {Causation as Influence},
  journal   = {The Journal of Philosophy},
  volume    = {97},
  number    = {4},
  pages     = {182--197},
  year      = {2000},
}

@book{pearl2000causality,
  author    = {Pearl, Judea},
  title     = {Causality: {Models}, Reasoning, and Inference},
  publisher = {Cambridge University Press},
  year      = {2000},
}

@inproceedings{pearl2001direct,
  author    = {Pearl, Judea},
  title     = {Direct and Indirect Effects},
  booktitle = {Proceedings of the Seventeenth Conference on Uncertainty
               in Artificial Intelligence ({UAI} 2001)},
  pages     = {411--420},
  publisher = {Morgan Kaufmann},
  year      = {2001},
}

@book{pearl2009causality,
  author    = {Pearl, Judea},
  title     = {Causality},
  edition   = {2},
  publisher = {Cambridge University Press},
  year      = {2009},
}

@book{salmon1984scientific,
  author    = {Salmon, Wesley C.},
  title     = {Scientific Explanation and the Causal Structure of the World},
  publisher = {Princeton University Press},
  year      = {1984},
}

@article{woodward2021explanatory,
  author    = {Woodward, James},
  title     = {Explanatory Autonomy: The Role of Proportionality,
               Stability, and Conditional Irrelevance},
  journal   = {Synthese},
  volume    = {198},
  pages     = {237--265},
  year      = {2021},
}

@article{halpern2016sufficient,
  author    = {Halpern, Joseph Y.},
  title     = {Sufficient Conditions for Causality to Be Transitive},
  journal   = {Philosophy of Science},
  volume    = {83},
  number    = {2},
  pages     = {213--226},
  year      = {2016},
}

@book{halpern2016actual,
  author    = {Halpern, Joseph Y.},
  title     = {Actual Causality},
  publisher = {MIT Press},
  year      = {2016},
}

@article{yablo1992mental,
  author    = {Yablo, Stephen},
  title     = {Mental Causation},
  journal   = {The Philosophical Review},
  volume    = {101},
  number    = {2},
  pages     = {245--280},
  year      = {1992},
}

@article{creel2020understanding,
  author    = {Creel, Kathleen A.},
  title     = {Transparency in Complex Computational Systems},
  journal   = {Philosophy of Science},
  volume    = {87},
  number    = {4},
  pages     = {568--589},
  year      = {2020},
}

@article{long2023introspective,
  author    = {Long, Robert},
  title     = {Introspective Capabilities in Large Language Models},
  journal   = {Journal of Consciousness Studies},
  volume    = {30},
  number    = {9--10},
  pages     = {143--153},
  year      = {2023},
  publisher = {Imprint Academic},
}

@article{kammerer2023forms,
  author    = {Kammerer, François and Frankish, Keith},
  title     = {What Forms Could Introspective Systems Take?
               {A} Research Programme},
  journal   = {Journal of Consciousness Studies},
  volume    = {30},
  number    = {9--10},
  pages     = {13--48},
  year      = {2023},
  publisher = {Imprint Academic},
}

@article{song2025privileged,
  title={Privileged Self-Access Matters for Introspection in {AI}},
  author={Song, Siyuan and Lederman, Harvey and Hu, Jennifer and Mahowald, Kyle},
  journal={arXiv preprint arXiv:2508.14802},
  year={2025}
}

@article{budding2025large,
  title={What Do Large Language Models Know? {T}acit Knowledge as a Potential Causal-Explanatory Structure},
  author={Budding, C{\'e}line},
  journal={Philosophy of Science},
  volume={92},
  number={4},
  pages={785--806},
  year={2025},
  publisher={Cambridge University Press}
}

@article{YetmanForthcoming-YETRIL-2,
	author = {Cameron Yetman},
	journal = {Ergo: An Open Access Journal of Philosophy},
	title = {Representation in Large Language Models},
	year = {forthcoming}
}

@article{beckmann2026mechanistic,
  title={Mechanistic indicators of understanding in large language models},
  author={Beckmann, Pierre and Queloz, Matthieu},
  journal={Philosophical Studies},
  pages={1--46},
  year={2026},
  publisher={Springer}
}

@article{milliere2024philosophical,
  title={A philosophical introduction to language models-part ii: The way forward},
  author={Milli{\`e}re, Rapha{\"e}l and Buckner, Cameron},
  journal={arXiv preprint arXiv:2405.03207},
  year={2024}
}

@article{goldstein2024doeschatgptmind,
      title={Does {C}hat{GPT} Have a Mind?}, 
      author={Simon Goldstein and Benjamin A. Levinstein},
      year={2026},
      journal={Philosophy of AI},
      doi={10.18716/ojs/phai/2026.11767},
      volume={2}, 
      pages={94-128}
}

@article{mollo2026vectorgroundingproblem,
      title={The Vector Grounding Problem}, 
      author={Coelho Mollo, Dimitri and Millière, Raphaël},
      year={2026},
      number={1}, 
      journal={Philosophy and the Mind Sciences}, 
      year={2026}, 
      month={Feb.}
}

@article{grindrod2026sparse,
  title={Sparse Auto-Encoders and Holism about Large Language Models},
  author={Grindrod, Jumbly},
  journal={arXiv preprint arXiv:2603.26207},
  year={2026}
}

@article{beckmann2025new,
  title={New horizons in machine understanding: explanatory and objectual understanding in deep learning video generation models},
  author={Beckmann, Pierre},
  journal={Synthese},
  volume={206},
  number={6},
  pages={285},
  year={2025},
  publisher={Springer}
}

@article{shafran2025decomposing,
  title={Decomposing mlp activations into interpretable features via semi-nonnegative matrix factorization},
  author={Shafran, Or and Geiger, Atticus and Geva, Mor},
  journal={arXiv preprint arXiv:2506.10920},
  year={2025}
}

@article{bricken2023monosemanticity,
  author  = {Bricken, Trenton and Templeton, Adamarie and Batson, Joshua and Chen, Brian and Jermyn, Adam and Conerly, Tom and Turner, Nick and Anil, Cem and Denison, Carson and Askell, Amanda and Lasenby, Robert and Wu, Yifan and Kravec, Shauna and Schiefer, Nicholas and Maxwell, Tim and Joseph, Nicholas and Hatfield-Dodds, Zac and Tamkin, Alex and Nguyen, Karina and McLean, Brayden and Burke, Josiah E. and Hume, Tristan and Carter, Shan and Henighan, Tom and Olah, Christopher},
  title   = {Towards Monosemanticity: Decomposing Language Models With Dictionary Learning},
  journal = {Transformer Circuits Thread},
  year    = {2023},
  url     = {https://transformer-circuits.pub/2023/monosemantic-features}
}

@inproceedings{
goldowsky-dill2025detecting,
title={Detecting Strategic Deception with Linear Probes},
author={Nicholas Goldowsky-Dill and Bilal Chughtai and Stefan Heimersheim and Marius Hobbhahn},
booktitle={Forty-second International Conference on Machine Learning},
year={2025},
url={https://openreview.net/forum?id=C5Jj3QKQav}
}

@inproceedings{ravfogel2020null,
  title={Null it out: Guarding protected attributes by iterative nullspace projection},
  author={Ravfogel, Shauli and Elazar, Yanai and Gonen, Hila and Twiton, Michael and Goldberg, Yoav},
  booktitle={Proceedings of the 58th annual meeting of the association for computational linguistics},
  pages={7237--7256},
  year={2020}
}

@misc{kadavath2022language,
  title         = {Language Models (Mostly) Know What They Know},
  author        = {Kadavath, Saurav and Conerly, Tom and Askell, Amanda and Henighan, Tom and Drain, Dawn and Perez, Ethan and Schiefer, Nicholas and Hatfield-Dodds, Zac and DasSarma, Nova and Tran-Johnson, Eli and Johnston, Scott and El-Showk, Sheer and Jones, Andy and Elhage, Nelson and Hume, Tristan and Chen, Anna and Bai, Yuntao and Bowman, Sam and Fort, Stanislav and Ganguli, Deep and Hernandez, Danny and Jacobson, Josh and Kernion, Jackson and Kravec, Shauna and Lovitt, Liane and Ndousse, Kamal and Olsson, Catherine and Ringer, Sam and Amodei, Dario and Brown, Tom and Clark, Jack and Joseph, Nicholas and Mann, Ben and McCandlish, Sam and Olah, Chris and Kaplan, Jared},
  year          = {2022},
  eprint        = {2207.05221},
  archivePrefix = {arXiv},
  primaryClass  = {cs.CL},
  url           = {https://arxiv.org/abs/2207.05221}
}

@article{lin2022teaching,
  title   = {Teaching Models to Express Their Uncertainty in Words},
  author  = {Lin, Stephanie and Hilton, Jacob and Evans, Owain},
  journal = {Transactions on Machine Learning Research},
  year    = {2022},
  url     = {https://arxiv.org/abs/2205.14334}
}

@inproceedings{binder2025looking,
  title     = {Looking Inward: Language Models Can Learn About Themselves by Introspection},
  author    = {Binder, Felix Jedidja and Chua, James and Korbak, Tomek and Sleight, Henry and Hughes, John and Long, Robert and Perez, Ethan and Turpin, Miles and Evans, Owain},
  booktitle = {International Conference on Learning Representations (ICLR)},
  year      = {2025},
  url       = {https://arxiv.org/abs/2410.13787}
}

@misc{lindsey2025emergent,
  title        = {Emergent Introspective Awareness in Large Language Models},
  author       = {Lindsey, Jack},
  year         = {2025},
  howpublished = {Transformer Circuits Thread},
  url          = {https://transformer-circuits.pub/2025/introspection/index.html}
}

@incollection{schwitzgebel2024introspection,
  author    = {Schwitzgebel, Eric},
  title     = {Introspection},
  booktitle = {The {Stanford} Encyclopedia of Philosophy},
  editor    = {Edward N. Zalta and Uri Nodelman},
  edition   = {{F}all 2024},
  year      = {2024},
  publisher = {Metaphysics Research Lab, Stanford University},
  url       = {https://plato.stanford.edu/archives/fall2024/entries/introspection/}
}

@misc{lederman2026emergent,
  title         = {Emergent Introspection in {AI} is Content-Agnostic},
  author        = {Lederman, Harvey and Mahowald, Kyle},
  year          = {2026},
  eprint        = {2603.05414},
  archivePrefix = {arXiv},
  primaryClass  = {cs.AI},
  url           = {https://arxiv.org/abs/2603.05414}
}

@article{comsa2025does,
  title={Does It Make Sense to Speak of Introspection in Large Language Models?},
  author={Comsa, Iulia M and Shanahan, Murray},
  journal={arXiv preprint arXiv:2506.05068},
  year={2025}
}

@article{Bechtel2009,
  title={Looking down, around, and up: Mechanistic explanation in psychology},
  author={Bechtel, William},
  journal={Philosophical Psychology},
  volume={22},
  number={5},
  pages={543--564},
  year={2009},
  publisher={Taylor \& Francis}
}

@article{holtzman2026prompting,
  title={Prompting as scientific inquiry},
  author={Holtzman, Ari and Tan, Chenhao},
  journal={Advances in Neural Information Processing Systems},
  volume={38},
  year={2026}
}

@article{dresow2025interdisciplinary,
  title={The interdisciplinary entanglement of characterization and explanation},
  author={Dresow, Max and Love, Alan C},
  journal={The British Journal for the Philosophy of Science},
  volume={76},
  number={3},
  pages={519--545},
  year={2025},
  publisher={The University of Chicago Press, Chicago, IL}
}

@article{prinzing2026role,
  title={The role of empirical evidence in philosophy},
  author={Prinzing, Michael},
  journal={Synthese},
  volume={207},
  number={5},
  pages={190},
  year={2026},
  publisher={Springer}
}

\end{document}